\documentclass[10pt,twocolumn,letterpaper]{article}

\usepackage[pagenumbers]{cvpr} 

\usepackage{pifont}
\usepackage{multirow}
\usepackage{booktabs}
\usepackage{xcolor}  
\usepackage{colortbl}
\definecolor{mygray}{gray}{.9}

\usepackage{arydshln}
\usepackage{makecell}
\usepackage{lipsum}
\usepackage{etoolbox}
\usepackage{subcaption}
\usepackage{float}
\usepackage{appendix}
\usepackage{multicol}
\usepackage{fontawesome}
\usepackage{marvosym}

\newcommand{\envelopeicon}{\textsuperscript{\Letter}}

\newcounter{appendixfigure}
\newcounter{appendixtable}
\renewcommand{\thefigure}{\arabic{figure}} 
\renewcommand{\thetable}{\arabic{table}}
\newcommand{\ifappendix}{%
  \setcounter{figure}{0} 
  \setcounter{table}{0}
  \setcounter{appendixfigure}{1} 
  \setcounter{appendixtable}{1}
  \renewcommand{\thefigure}{\Alph{section}.\arabic{appendixfigure}} 
  \renewcommand{\thetable}{\Alph{section}.\arabic{appendixtable}}
}
\definecolor{cvprblue}{rgb}{0.21,0.49,0.74}
\usepackage[pagebackref,breaklinks,colorlinks,allcolors=cvprblue]{hyperref}

\definecolor{Grey}{RGB}{119,119,119}
\newcommand{\demph}[1]{\textcolor{Grey}{#1}}

\newcommand{\CUT}[1]

\def\paperID{1899} 
\def\confName{CVPR}
\def\confYear{2025}

\title{\textit{DistinctAD}: Distinctive Audio Description Generation in Contexts}

\author{Bo Fang$^{1}$, Wenhao Wu$^{2,3}$, Qiangqiang Wu$^{1}$, Yuxin Song$^{2}$, Antoni B. Chan$^{1}\envelopeicon$\\
$^{1}$ Department of Computer Science, City University of Hong Kong\\
$^{2}$ Baidu Inc. \qquad $^{3}$ The University of Sydney \\
{\tt\small \{bofang6-c,qiangqwu2-c\}@my.cityu.edu.hk, wenhao.wu@sydney.edu.au}\\
{\tt\small songyuxin02@baidu.com, abchan@cityu.edu.hk}
}

\begin{document}
\maketitle
\begin{abstract}
Audio Descriptions (ADs) aim to provide a narration of a movie in text form, describing non-dialogue-related narratives, such as characters, actions, or scene establishment. Automatic generation of ADs
remains challenging due to: i) the domain gap between movie-AD data and existing data used to train vision-language models,
 and ii) the issue of contextual redundancy arising from highly similar neighboring visual clips in a long movie.
In this work, we propose \textbf{DistinctAD}, a novel two-stage framework for generating ADs that emphasize distinctiveness to produce better narratives.
%
To address the domain gap, we introduce a CLIP-AD adaptation strategy that does not require additional AD corpora, enabling more effective alignment between movie and AD modalities at both global and fine-grained levels. 
In Stage-II, DistinctAD incorporates two key innovations: (i) a Contextual Expectation-Maximization Attention (EMA) module that reduces redundancy by extracting common bases from consecutive video clips, and (ii) an explicit distinctive word prediction loss that filters out repeated words in the context, ensuring the prediction of unique terms specific to the current AD. 
Comprehensive evaluations on MAD-Eval, CMD-AD, and TV-AD benchmarks demonstrate the superiority of DistinctAD, with the model consistently outperforming baselines, particularly in Recall@k/N, highlighting its effectiveness in producing high-quality, distinctive ADs.
\end{abstract}    
\section{Introduction}
\label{sec:intro}

\CUT{
\begin{quote}
\setlength{\parindent}{0pt}
\itshape
Just because a man lacks the use of his eyes doesn't mean he lacks vision. 
\vspace{-1.5em}
\begin{flushright}
Stevie Wonder
\end{flushright}
\end{quote}
\vspace{-1.5em}
\noindent\hrulefill
\vspace{0.5em}
}

\begin{figure}[h]
  \centering
  \includegraphics[width=1.\linewidth]{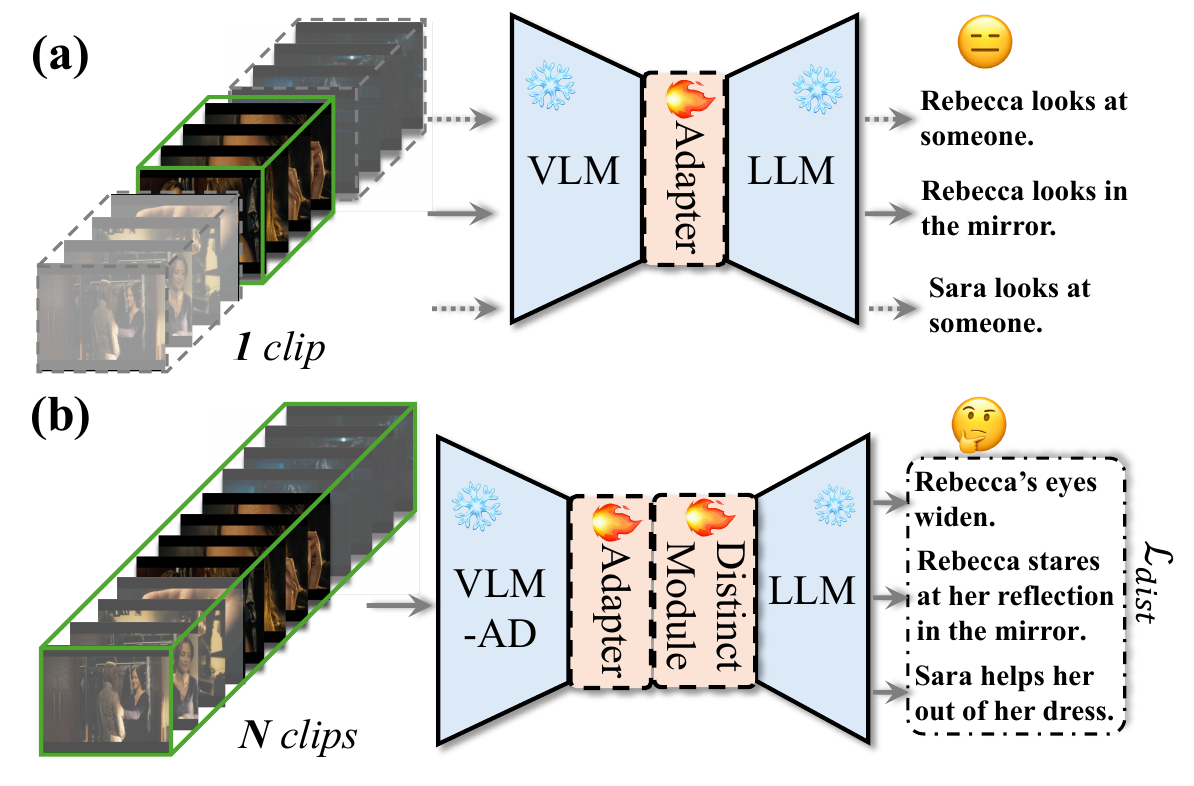}
   \vspace{-20pt}
   \caption{
   \textbf{(a)} Previous methods approach the AD task similar to video captioning, using only a single video clip as input, which leads to repetitive ADs due to highly similar neighboring clips.  
   \textbf{(b)} Our DistinctAD method generates distinctive ADs across $N$ consecutive clips, with three key innovations: VLM-AD adaptation, the Distinct Module, and explicit distinctive words prediction.
   }
   \label{fig:motivation}
   \vspace{-6mm}
\end{figure}

\noindent
\textit{Audio description (AD)}~\cite{snyder2005audio,fryer2016introduction} is a crucial accessibility service that provides verbal narration of visual elements in media content for individuals who are blind or have low vision. 
By offering succinct and vivid descriptions, ADs enable visually impaired audiences to fully comprehend and engage with non-dialogue-related narratives, \eg., characters, facial expressions, non-verbal actions, or scene establishment.
Recent studies also show ADs' value for sighted viewers in supporting eye-free activities and facilitating child language development~\cite{lewis2023deep,perego2016gains},
reinforcing its pivotal role in fostering inclusivity by bridging the perceptual gap between visual and non-visual elements. 
Crafting ADs requires careful attention to timing, language, and context to integrate smoothly with dialogue~\cite{han2023autoad_ii}. However, despite the availability of advanced AD platforms~\cite{pavel2020rescribe,branje2012livedescribe}, human-annotated methods are costly and difficult to scale, highlighting the need for automated generation systems, especially with the rise of user-generated content.


Advancements in Vision-Language Models (VLMs) and Large-Language Models (LLMs) have led to growing interest in automatic AD generation for media.
Current approaches fall into two categories: (i) using powerful proprietary models like GPT-4~\cite{achiam2023gpt4} or GPT-4V~\cite{achiam2023gpt4v} in a training-free manner~\cite{xie2024autoad_zero,zhang2024mmnarrator,lin2023mm-vid,chu2024LLM_AD}, and (ii) fine-tuning open-source VLM components, such as visual-text adapters~\cite{alayrac2022flamingo,li2023blip-2,liu2024llava}, for AD tasks~\cite{han2023autoad,han2023autoad_ii,han2024autoad_iii,wang2024contextual_AD,lin2024movieseq,raajesh2024micap}.
Both approaches have limitations: (i) Training-free methods often perform poorly and suffer from hallucinations due to the unique nature of AD (\eg, character names and narrative coherence), which differs from the common text data LLMs are trained on.
(ii) Fine-tuning methods generally perform better but are still limited by insufficient data to fully adapt to the movie-AD domain and face the \textit{context-repetition} issue.


%
Unlike video captioning~\cite{lin2022swinbert,seo2022end}, ADs are generated on consecutive intervals (visual clips) throughout long videos~\cite{soldan2022MAD}, \eg, movies.
The \textit{context-repetition} issue arises when models produce repetitive or similar descriptions for consecutive visual clips, especially when using prior ADs as prompts~\cite{han2023autoad,wang2024contextual_AD}. This occurs because sequential clips often comprise redundant scenes or characters (and therein redundant  visual features), leading models 
that only use the current visual clip to repeat the same information from the past, as shown in \cref{fig:motivation}.
However, audiences are more interested in the unique and distinct events of the current clip, rather than the common elements from the previous one.

In this paper, we propose \textbf{DistinctAD}, a two-stage framework for generating distinctive ADs within contexts. 
Given the domain gap between the movie-AD and VLM training data, we first bridge this gap in Stage-I by adapting VLMs, such as CLIP~\cite{radford2021CLIP}, to the movie-AD domain. Our adaptation strategy is inspired by a key observation (see Appendix \S\ref{sec:supp_reconstruct_ad}): AD sentences encoded by the CLIP text encoder can be effectively reconstructed using simple LLMs like GPT-2~\cite{radford2019GPT-2} with minimal fine-tuning, whereas AD reconstructions using CLIP visual features from the corresponding clips are often of poor quality.
\emph{This suggests that while CLIP's multi-modal embedding space is rich enough to represent AD information, its visual encoder is insufficient for extracting it.}
%
To mitigate this domain gap,
we 
adapt the CLIP vision encoder to better align with the frozen CLIP text encoder using existing paired video-AD data. 
The alignment involves global matching at video-sentence level, similar to CLIP pre-training.
A challenge arises because video clips are labeled with whole ADs, and words may not appear in every frame but must be aggregated over frames. Therefore, we propose fine-grained matching at frame-word level for this \emph{multiple-instance setting}.

For Stage-II, we propose a novel distinctive AD narrating pipeline based on the Expectation-Maximization Attention (EMA)~\cite{dempster1977maximum} algorithm, which has 
demonstrated its efficacy 
in tasks such as semantic segmentation~\cite{li2019ema}, video object segmentation~\cite{lin2022swem}, and text-video retrieval~\cite{jin2022expectation}.
Differently, we apply EMA to contextual clips from long videos, which often exhibit high redundancy due to recurring scenes or characters. 
By extracting common bases from contextual information, DistinctAD reduces redundancy and generates compact, discriminative representations 
that enable the LLM decoder to produce more distinctive ADs.
To further emphasize distinctiveness explicitly, we introduce a distinctive word prediction loss that filters out words that repeatedly appear in contexts, ensuring that the LLM decoder focuses on predicting unique words specific to the current AD. 
With these two designs, DistinctAD produces contextually distinctive and engaging ADs 
that can provide better narratives for the audience.

In summary, our contributions are three-fold:
\begin{itemize}
\item We propose a CLIP-AD adaptation strategy tailored to movie-AD data, addressing the misalignment issue caused by the domain gap. Our adapted vision encoder can be seamlessly integrated into existing CLIP-based AD methods and stands to benefit from future improvements as more AD data becomes available. 
\item We introduce DistinctAD, which incorporates a Contextual EMA module and a distinctive word prediction loss, significantly enhancing the generation of distinctive ADs from consecutive visual clips with similar contexts. 
%
\item Comprehensive evaluations on MAD-Eval~\cite{han2023autoad}, CMD-AD~\cite{han2024autoad_iii}, and TV-AD~\cite{xie2024autoad_zero} highlight DistinctAD's superiority. Our outstanding performance in Recall@k/N demonstrates its effectiveness in generating high-quality ADs with both distinctiveness and technical excellence.
\end{itemize}

\section{Related Work}
\label{sec:literature}

\noindent
\textbf{Dense video captioning.}
A task closely related to AD is \textit{dense} video captioning~\cite{krishna2017dense}, which extends traditional video captioning~\cite{lin2022swinbert,seo2022end,luo2020univl,sharma2018conceptual} by both generating a single caption for trimmed video segments as well as  detecting and describing multiple events with grounded timestamps.
Initial dense video captioning utilize a 2-stage pipeline~\cite{iashin2020better,iashin2020multi,wang2018bidirectional,wang2020event} by firstly performing localization and then describing events.
Recent works~\cite{wang2018bidirectional,wang2021end,zhou2018end,chen2021towards,deng2021sketch,li2018jointly,mun2019streamlined,rahman2019watch,shen2017weakly,shi2019dense,yang2023vid2seq} focus on training localization and captioning modules in an end-to-end manner to enhance inter-event associations.
In contrast to these works, AD generation specifically aims to narrate a coherent story, maintain character-awareness, and complement the audio track without interfering with existing dialogue.

\noindent
\textbf{AD generation.}
ADs narrate key visual elements in extended videos, enabling blind and visually-impaired audiences to appreciate films, TV series, \etc. 
Early AD systems relied heavily on specialized authoring tools~\cite{branje2012livedescribe} and skilled human contributors. Platforms like Rescribe~\cite{pavel2020rescribe} and LiveDescribe~\cite{branje2012livedescribe} have facilitated faster and more accurate AD creation; however, these methods are costly and do not scale efficiently for large volumes of visual content.
Recent efforts have developed audio segmentation and transcription systems~\cite{bain2023whisperx,bredin2021end_speaker_seg,bredin2020pyannote} to create high-quality video datasets with temporally aligned ADs~\cite{rohrbach2015dataset,rohrbach2017movie,soldan2022MAD,torabi2015using}, advancing automatic AD research.

In general, current automatic AD generation systems can be categorized into two approaches: training-free and partial-fine-tuning. 
\textit{Training-free} methods~\cite{lin2023mm-vid} 
generate
ADs by leveraging 
proprietary models like GPT-4~\cite{achiam2023gpt4} and GPT-4V~\cite{achiam2023gpt4v}. 
MM-Narrator~\cite{zhang2024mmnarrator} enhances AD performance by multi-model in-context learning with memories.
LLM-AD~\cite{chu2024LLM_AD} and AutoAD-Zero~\cite{xie2024autoad_zero} use prompts comprising visual frames with textual character names and colorful circles~\cite{shtedritski2023redcircle}, enabling character-centric AD generation. 
However, training-free AD methods often suffer from high evaluation costs at scale and relatively poor performance due to domain-specific challenges and LLM hallucinations. 
 \textit{Partial-fine-tuning} methods~\cite{han2023autoad,han2023autoad_ii,han2024autoad_iii,lin2024movieseq,wang2024contextual_AD}, as well as our DistinctAD, only fine-tune a lightweight adapter~\cite{alayrac2022flamingo,li2023blip-2} between the pre-trained vision and text encoders.
A representative example is the AutoAD series~\cite{han2023autoad,han2023autoad_ii,han2024autoad_iii}, which builds automatic AD systems and enriches them with character-aware prompts within different vision-language frameworks.
However, previous studies tend to focus on constructing more accurate external character banks, whereas treating AD generation similarly to video captioning, overlooks AD's 
unique sequential structure of video clips.
In contrast, our method emphasizes understanding the visual content within its temporal context, leading to more distinctive AD generation.

\noindent
\textbf{Distinctive captioning} 
in images aims to articulate unique details that can help distinguishing targets from others.
An intuitive way to promoting distinctiveness is through contrastive learning~\cite{dai2017contrastive,liu2018show,luo2018discriminability,vered2019joint}, where generated captions are encouraged to align more closely with target images rather than distractors.
In~\cite{chen2018groupcap,wang2020compare,wang2022distinctive,wang2021group}, group-based distinctive attention is introduced to capture distinctiveness by comparing sets of similar images and re-weighting specific caption words.
A recent closely related field is difference captioning~\cite{yao2022image_difference,li2023mimic,li2023otter}, which aims to describe differences between a single pair of images. VisDiff~\cite{dunlap2024describing} scales difference captioning to sets containing thousands of images with natural language.
Our work differs from these distinctive captioning works in that we are the first to explore distinctiveness across dense, consecutive clips within hours-long movies, thereby generating 
ADs with better narrative.
\section{Method}
\label{sec:method}


This section outlines the DistinctAD pipeline, consisting of \textit{two stages} for AD generation. 

\subsection{Stage-I: CLIP-AD Adaptation}
\label{subsec:stage_1}

\begin{figure}[t]
  \centering
   \includegraphics[width=1.0\linewidth]{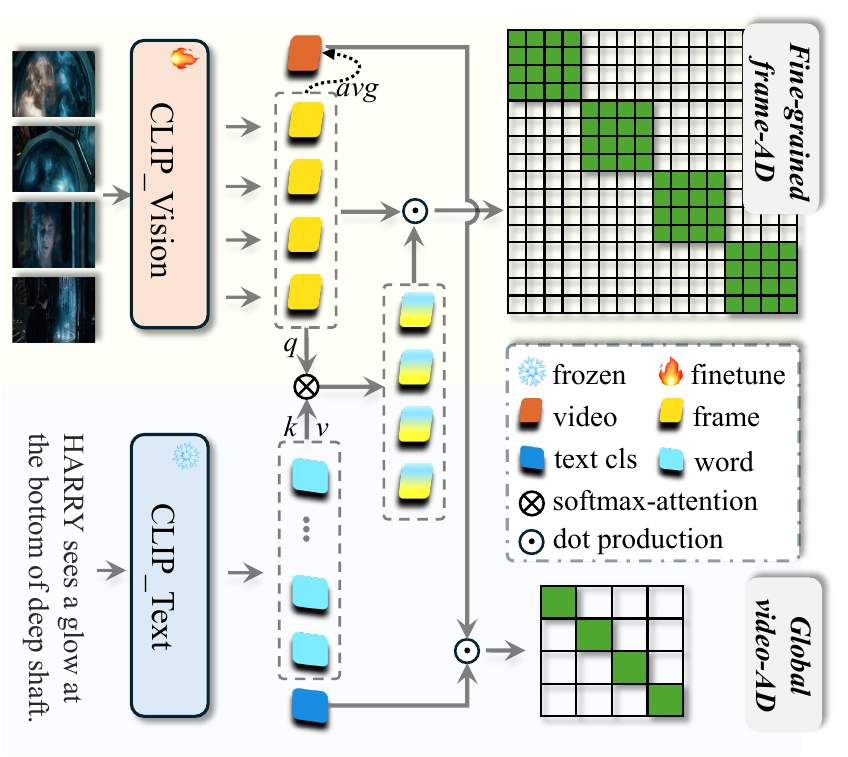}
   \vspace{-25pt}
   \caption{Illustration of Stage-I: CLIP-AD Adaptation. This process involves adapting the CLIP vision encoder to specific movie-AD data through global-level video-AD matching (bottom right) and fine-grained frame-AD matching 
   (top right).}
   \label{fig:stage_i_fig}
   \vspace{-4mm}
\end{figure}

AD and the visual content it describes exhibit a significant domain gap compared to typical large-scale web data. 
This gap often causes misalignment in current partial-fine-tuning techniques.
%
Previous studies~\cite{han2023autoad,han2023autoad_ii,wang2024contextual_AD} try alleviating this problem by pre-training LLMs on text-only AD corpus, \eg AudioVault$^1$\footnote{$^1$ \href{https://audiovault.net}{\texttt{https://audiovault.net}}}.
However, misalignment persists at the initial stage of vision encoding, which is often neglected.

Inspired by our findings that AD sentences
encoded by the CLIP text encoder 
can be effectively recovered using the GPT-2 language model (see \S\ref{sec:intro} and Appendix \S\ref{sec:supp_reconstruct_ad}),  
we identify that the primary issue of misalignment is caused by the CLIP vision encoder, 
\ie, 
the discrepancy between visual embeddings and AD embeddings within the joint CLIP feature space. 
To address this, we propose adapting the CLIP vision encoder to the specific AD domain.
%
However, due to the unique multiple-instance learning setting for video-AD pairs (see \S\ref{sec:intro}), we consider both global matching and fine-grained frame-word matching in our adaptation method. 

\noindent 
\textbf{Global video-AD matching.} A straightforward strategy involves adopting classical CLIP-style fine-tuning with video-AD pairs in large batches~\cite{luo2022clip4clip}.
Formally, let video clip $\mathbf{V}_i = [\mathbf{f}_i^1; \cdots; \mathbf{f}_i^n]\in\mathbb{R}^{n\times C}$ be a collection of $n$ frame embeddings,  and corresponding AD
$\mathbf{T}_i=[\mathbf{w}_i^0; \mathbf{w}_i^1; \cdots; \mathbf{w}_i^m] \in \mathbb{R}^{(m+1)\times C}$ be a collection of $m$ word embeddings  ($\mathbf{w}_i^j$) and the \texttt{[CLS]} token (denoted as $\mathbf{w}_i^0$), where $C$ is the number of channels in the embedding space. 
We obtain the \emph{global} 
video-level representation by averaging all frame embeddings in $\mathbf{V}_i$ using mean pooling: $\mathbf{v}_i = \frac{1}{n}\sum_{j=1}^n  \mathbf{f}_i^j$. 
%
Following the standard CLIP, we use the  \texttt{[CLS]} token as the \emph{global} textual AD representation $\mathbf{t}_i = \mathbf{w}_i^0$.
The \textbf{global video$\rightarrow$AD matching} is performed 
by maximizing the sum of the main diagonal of a $B\times B$ similarity matrix, using the contrastive loss:
\vspace{-0.5em}
\begin{equation}
    \mathcal{L}_{v\rightarrow AD} = -\frac{1}{B}\sum_{i=1}^{B} \mathrm{log} \frac{\exp(sim(\mathbf{v}_i, \mathbf{t}_i))}{\sum_{j=1}^{B} 
    \exp (sim(\mathbf{v}_i, \mathbf{t}_j))},
    \label{eq:v2t}
\end{equation}
where $B$ is the batch size, and the similarity function $sim(\cdot)$ is the vector inner product.
This process is illustrated in the bottom right of \cref{fig:stage_i_fig}.
Similarly, we drive the AD$\rightarrow$video loss $\mathcal{L}_{AD\rightarrow v}$ by maximizing the sum of the secondary diagonal (i.e., swapping the $i$ and $j$ indices in (\ref{eq:v2t})).
The final global-level contrastive loss is then the sum of the losses in both directions $\mathcal{L}_g=\mathcal{L}_{v\rightarrow AD}+\mathcal{L}_{AD\rightarrow v}$.

\begin{figure*}[h]
  \centering
   \includegraphics[width=0.78\linewidth]{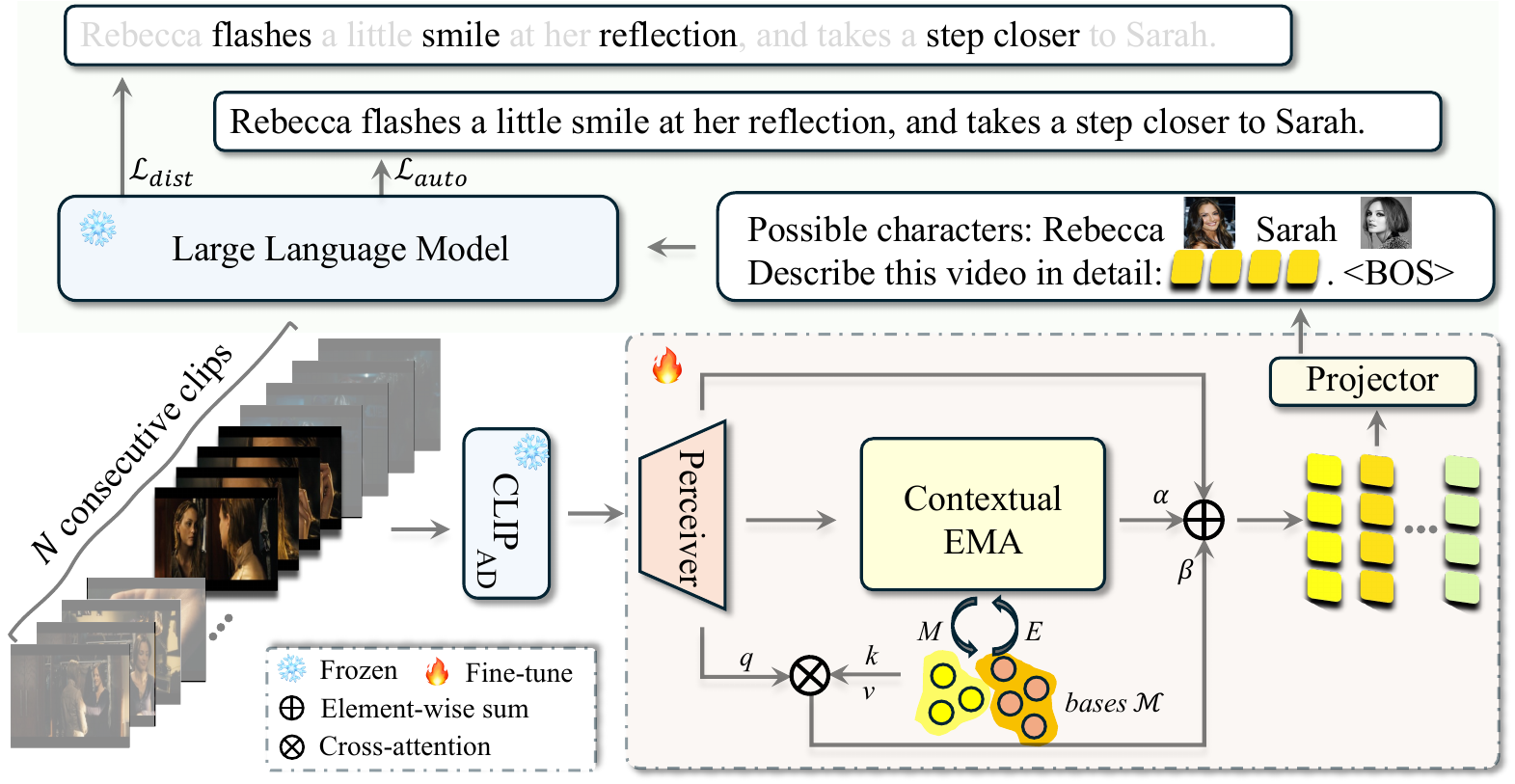}
   \vspace{-8pt}
   \caption{Pipeline of Stage-II: Distinctive AD Narration. 
   Stage-II processes $N$ consecutive video clips using the CLIP$_{\mathrm{AD}}$ vision encoder from Stage-I.
   We generate contextual-distinctive ADs by two key innovations: \textit{i)} a Contextual EMA module to learn compact and discriminative visual representations for improved prompting of LLMs; \textit{ii)} an extra distinctive word 
   loss for predicting AD-specific terms.}
   \label{fig:stage_ii_fig}
   \vspace{-4mm}
\end{figure*}

\noindent
\textbf{Fine-grained frame-AD matching.}
Matching global video to AD sentence \texttt{[CLS]} (and vice versa) aids in joint feature space alignment. 
However, this alignment is insufficient for effective adaptation due to the specific 
\emph{multiple-instance} setting of ADs, where only some words may correspond to a particular frame, but all words will have correspondence in aggregate.
%
Thus, we propose a fine-grained matching loss at the frame-level to address this issue.

%
Formally, 
given the frame embeddings $\mathbf{V}_i$
and the word embeddings $\mathbf{T}'_i = [\mathbf{w}_i^1;\cdots;\mathbf{w}_i^m] \in \mathbb{R}^{m\times C}$, 
we calculate the weights of all words attending to each frame via softmax attention, taking $\mathbf{V}_i$ as the query and $\mathbf{T}'_i$ as the key.
By then multiplying these attention weights by the value  $\mathbf{T}'_i$, we obtain 
a frame-aware AD representation $\tilde{\mathbf{T}}_i \in \mathbb{R}^{n\times C}$:
\vspace{-0.5em}
\begin{equation}
    \tilde{\mathbf{T}}_i = \mathrm{Softmax}(\mathbf{V}_i {\mathbf{T}'_i}^T / \tau) \mathbf{T}'_i
    = [\tilde{\mathbf{t}}_i^1;\cdots;
    \tilde{\mathbf{t}}_i^n],
    \label{eq:softmax-attention}
\end{equation}
where for each frame, the words embeddings that are most similar to the frame-level visual feature have been aggregated (via softmax attention).
The temperature parameter  $\tau$ controls the aggregation process, where smaller $\tau$ incorporates more textual information.

The goal of the fine-grained matching is to pull a 
frame visual feature $\mathbf{f} \in \mathbf{V}_i$ closer to 
the frame-aware AD representations $\tilde{\mathbf{t}} \in \tilde{\mathbf{T}}_i$ 
in (\ref{eq:softmax-attention}), corresponding to the positive set.
To achieve this, we define the negative set $\tilde{\mathbf{T}}_{neg}$ as 
the frame-aware AD embeddings
generated from other video clips (in the batch), 
and then use a Multi-Instance Loss~\cite{miech2020mutliins}, 
\vspace{-0.5em}
\begin{equation}
    \mathcal{L}_{f} = 
    -\frac{1}{B}\sum_{i=1}^{B}
    \mathrm{log}
    \frac{\sum_{\tilde{\mathbf{t}}\in \tilde{\mathbf{T}}_i} \exp (sim(\mathbf{f}, \tilde{\mathbf{t}}))}
    {\sum_{\tilde{\mathbf{t}}_* \in \{ \tilde{\mathbf{T}}_i \cup \tilde{\mathbf{T}}_{neg} \} } \mathrm{exp} (sim(\mathbf{f}, \tilde{\mathbf{t}}_*))}, 
\end{equation}
where $\mathbf{f} \in \mathbf{V}_i$ is a sampled frame from $\mathbf{V}_i$.
This process is illustrated in the top right of \cref{fig:stage_i_fig}.

\noindent
\textbf{Summary for Stage-I.}
The final objective for Stage-I is to minimize the sum of \textbf{global} and \textbf{fine-grained} aligning losses, balanced by a trade-off coefficient, $\mathcal{L}_{\mathrm I}=\gamma \mathcal{L}_g + (1-\gamma)\mathcal{L}_f$.
Note that during this adaptation process, the CLIP-Text encoder model remains frozen, and only the CLIP-Vision encoder is fine-tuned. Our fine-grained frame-AD matching is entirely parameter-free, as only the vision encoder will be utilized in the subsequent stage.

\subsection{Stage-II: Distinctive AD Narration}
\label{subsec:stage_2}

The motivation for generating \textit{distinctive} ADs stems from the observation that LLM often produce repetitive descriptions for adjacent clips~\cite{xie2024autoad_zero,zhang2024mmnarrator,raajesh2024micap}.
Despite improved character recognition, the visual representation itself is not discriminative among neighboring (contextual) clips, leading to uninteresting ADs.
Our goal is to create contextual-distinctive ADs that highlight current differences.
We hypothesize, as verified in Appendix \S\ref{sec:supp_similar_ctx}
, that sequential clips from a long video often share redundant scenes or characters, leading similar visual features in contexts.
Thus, we propose Stage-II: distinctive AD narration.

As shown in \cref{fig:stage_ii_fig}, we prepare $N$ consecutive video clips (to be AD-described) $\{\mathbf{x}_1, \mathbf{x}_2, \cdots, \mathbf{x}_N\}$, each containing $T$ uniformly sampled frames $\{\mathcal{F}_1, \mathcal{F}_2, \cdots, \mathcal{F}_T\}$.
Following~\cite{han2023autoad,han2023autoad_ii,wang2024contextual_AD}, we employ a learnable Perceiver adapter~\cite{alayrac2022flamingo} to resample $T'$ prompt vectors for the $T$ frame embeddings encoded by our Stage-I vision encoder, CLIP$_{\mathrm{AD}}$. 
This process is formulated as:
\vspace{-0.5em}
\begin{align}
    \centering
    \mathbf{h}_{\mathbf{x}_i} &= \mathrm{Perceiver}(\{\mathbf{f}_1, \mathbf{f}_2, \cdots, \mathbf{f}_T\}) \in \mathbb{R}^{T'\times C},
    \\
    \mathbf{f}_i &= \mathrm{CLIP}_{\mathrm{AD}}(\mathcal{F}_i).
\end{align}
%
We then introduce the Contextual EMA to capture compact, discriminative visual features for distinctive AD generation.

\noindent
\textbf{Contextual EMA.}
Expectation-Maximization Attention (EMA)~\cite{li2019ema} integrates the attention mechanism~\cite{wang2018non-local} into the classical EM~\cite{dempster1977maximum} algorithm, which comprises three steps to estimate a more compact set of bases: \textit{Responsibility
Estimation} (RE), \textit{Likelihood Maximization} (LM), and \textit{Data Re-estimation} (DR). 
Inspired by this, we propose \textbf{Contextual EMA} to perform EMA on frames from $N$ contextual clips, 
aiming to eliminate redundancy, learn compact representations, and explore distinctiveness.

Let $\mathcal{H}=\{\mathbf{h}_{\mathbf{x}_i}\}_{i=1}^{N} \in \mathbb{R}^{N\times T'\times C}$ represent $N$ clip vectors from the Perceiver, and $\mathcal{M}=\{\mathbb{\mu}_k\}_{k=1}^K \in \mathbb{R}^{K\times C}$ denote the randomly initialized base features, where $C, K$ indicate the number of channel and bases.
The RE step estimates the hidden variable $\mathcal{Z}=\{z_{nk}\}_{n=1,k=1}^{N\times T',K}$, where the responsibility $z_{nk}$ represents the probability of the \textit{n}-th \textit{frame} belonging to the \textit{k}-th base:
\vspace{-0.5em}
\begin{equation}
    z_{nk} = \frac{\exp(\mathbf{h}_n\mu_k^T/\tau)}
    {\sum_{j=1}^K \exp(\mathbf{h}_n\mu_j^T/\tau)},
    \label{eq:z_nk}
\end{equation}
where $\tau$ determines the shape (peakiness) of distribution $\mathcal{Z}$.
Then, the LM step updates the bases $\mathcal{M}$ by applying the weighted average
on input $\mathcal{H}$, formulating the $k$-th base as:
\vspace{-0.5em}
\begin{equation}
    \mu_k = \frac{\sum_{n=1}^{N\times T'} z_{nk}\mathbf{h}_n }{\sum_{n=1}^{N\times T'} z_{nk}}.
    \label{eq:mu_k}
\end{equation}
The RE (E-step) and LM (M-step) are iteratively performed $R$ times until convergence. 
Notably, since bases number $K$ is much smaller than the embedding number $N\times T'$, we employ DR to reconstruct a compact version of $\mathcal{H}$ through:
\vspace{-0.5em}
\begin{equation}
    \widehat{\mathcal{H}} \approx \mathcal{Z}\mathcal{M}.
\end{equation}
Here, $\widehat{\mathcal{H}} \in \mathbb{R}^{N\times T'\times C}$ retains the same shape as $\mathcal{H}$. We combine $\mathcal{H}$ and $\widehat{\mathcal{H}}$ element-wise with a hyperparameter $\alpha$.

To enhance representation distinctiveness, we introduce an additional branch using cross-attention between raw $\mathcal{H}$ (query) and bases $\mathcal{M}$ (key and value), formulated as:
\vspace{-0.5em}
\begin{equation}
    \widetilde{\mathcal{H}} = \mathrm{CrossAttention}(\mathcal{H}, \mathcal{M}),
    \label{eq:x_attn}
\end{equation}
where $\widetilde{\mathcal{H}}\in \mathbb{R}^{N\times T'\times C}$.
Linear layers projecting queries, keys, and values are omitted in \cref{fig:stage_ii_fig} (see Appendix \S\ref{sec:supp_x_attn} for details).
Through (\ref{eq:x_attn}), 
we process 
the distributions of $\mathcal{H}$ to attend on specific and informative bases, with improved linear separability (see \cref{fig:tsne_vis}).
We combine $\mathcal{H}, \widehat{\mathcal{H}}, 
\widetilde{\mathcal{H}}$ elementwise around Contextual EMA to construct the final refined visual features. These features are then projected into the LLM embedding space using a single-layer projector:
\vspace{-0.5em}
\begin{equation}
    \mathcal{H}_{sum} = \mathrm{Proj}(\mathcal{H}+\alpha\widehat{\mathcal{H}}+\beta\widetilde{\mathcal{H}}).
\end{equation}

\noindent
\textbf{Interleaved prompt as LLM's input.} 
Following previous studies~\cite{han2023autoad_ii,xie2024autoad_zero,wang2024contextual_AD}, we build our interleaved \texttt{prompt} enriched with character information, 
(see \cref{fig:stage_ii_fig}). 
To answer the ``who is who" question when more than two characters are present, the corresponding actors' portrait images are projected as face tokens for reasoning.
The \texttt{<BOS>} tag appended at the end indicates the beginning of AD generation.

\noindent
\textbf{Distinctive words highlighting.}
Our goal is to query a frozen LLM for AD generation using a vision-conditioned prompt. The typical supervision employs the commonly used auto-regressive loss function:
\begin{equation}
    \mathcal{L}_{auto} = -\sum_n \log P_{\theta}(w_n|\mathrm{prompt};w_{<n}),
    \label{eq:auto_regress}
\end{equation}
where $w_n$ is the $n$-th token from the target AD. 
However, $\mathcal{L}_{auto}$ does not emphasize the distinctiveness specific to the current AD, which is our focus. To address this, we propose a distinctive word set $w_d$, created by filtering out duplicates, such as character names, prepositions, and pronouns, from the $N$ context ADs of the target AD. During training, we explicitly encourage the LLM to predict the distinctive words in $w_d$ by optimizing the distinctive loss $\mathcal{L}_{dist}$:

\vspace{-0.5em}
\begin{equation}
    \mathcal{L}_{dist} = -\sum_{n=1}^{N}\sum_{i=1}^{u}\log P_{\theta}(w_n=w_d^i|\mathrm{prompt}, w_{<n}),
    \label{eq:loss_dist}
\end{equation}
where $w_d^i$ denotes the $i$-th distinctive word in $w_d$ and $u$ is the size of the set.
The final complete loss function for Stage-II is: $\mathcal{L}_{\mathrm{II}}=\mathcal{L}_{auto}+\mathcal{L}_{dist}$.



\section{Experiments}
\label{sec:experiment}

\begin{table*}
  \centering
  \scalebox{0.9}{
  \begin{tabular}{lccccccc}
    \toprule
    Method & Pub.  & VLM & LLM & ROUGE-L & CIDEr & SPICE & R@5/16 \\
    \hline
    \rowcolor{blue!5} \textit{Training-free} & & & & & & & \\
    \hdashline[0.5pt/5pt]
    VLog~\cite{VLog} & -  & - & GPT-4 & 7.5 & 1.3 & 2.1 & 42.3 \\
    MM-Vid~\cite{lin2023mm-vid} & ArXiv'23 &  GPT-4V & - & 9.8 & 6.1 & 3.8 & 46.1 \\
    MM-Narrator~\cite{zhang2024mmnarrator} & CVPR'24 & CLIP-L14 & GPT-4 & 13.4 & 13.9 & 5.2 & 49.0 \\
    LLM-AD~\cite{chu2024LLM_AD} & ArXiv'24 & GPT-4V & - & 13.5 & 20.5 & - & - \\
    AutoAD-Zero~\cite{xie2024autoad_zero} & ACCV'24 & VideoLLaMA2-7B & LLaMA3-8B & - & 22.4 & - & - \\
    \hline
    \rowcolor{orange!5}\textit{Partial-fine-tuning} & & & & & & & \\
    \hdashline[0.5pt/5pt]
    ClipCap~\cite{mokady2021clipcap} & ArXiv'21 & CLIP-B32 & GPT-2 & 8.5 & 4.4 & 1.1 & - \\
    CapDec~\cite{nukrai2022CapDec} & ArXiv'22 & - & - & 8.2 & 6.7 & 1.4 & - \\
    AutoAD-I~\cite{han2023autoad} & CVPR'23 & CLIP-B32 & GPT-2 & 11.9 & 14.3 & 4.4 & 42.1 \\
    AutoAD-II~\cite{han2023autoad_ii}& ICCV'23 & CLIP-B32 & GPT-2 & 13.4 & 19.5 & - & 50.8 \\
    AutoAD-III~\cite{han2024autoad_iii} & CVPR'24 & EVA-CLIP & OPT-2.7B & - & 22.8 & - & 52.0 \\
    AutoAD-III~\cite{han2024autoad_iii} &CVPR'24 &  EVA-CLIP & LLaMA2-7B & - & 24.0 & - & 52.8 \\
    MovieSeq~\cite{lin2024movieseq} & ECCV'24 & CLIP-B16 & LLaMA2-7B$^{*}$ & 15.5 & 24.4 & 7.0 & 51.6 \\
    \rowcolor{gray!10} \textbf{DistinctAD} (Ours) &  & CLIP-B32 & GPT-2 & 15.4 & 24.5 & 6.7 & 49.8 \\
    \rowcolor{gray!10} \textbf{DistinctAD} (Ours) &  & CLIP$_{\mathrm{AD}}$-B32 & GPT-2 & 16.4 & 25.5 & 7.4 & 51.7 \\
    \rowcolor{gray!10} \textbf{DistinctAD} (Ours) &  & CLIP$_{\mathrm{AD}}$-B16 & LLaMA2-7B & 17.2 & 27.0 & 8.2 & 55.6 \\
    \rowcolor{gray!10} \textbf{DistinctAD} (Ours) &  & CLIP$_{\mathrm{AD}}$-B16 & LLaMA3-8B & \textbf{17.6} & \textbf{27.3} & \textbf{8.3} & \textbf{56.0} \\
    \bottomrule
  \end{tabular}}
  \vspace{-2mm}
  \caption{
  Comparisons of AD performance on the MAD-Eval benchmark.
  $^{*}$ indicates fine-tuning LLaMA2-7B model with LoRA~\cite{hu2021lora}. 
  CLIP$_{\mathrm{AD}}$ is our
CLIP vision encoder adapted using our Stage-I strategy.}
  \label{tab:mad-sota}
  \vspace{-3mm}
\end{table*}

\begin{table}
  \centering
  \setlength{\tabcolsep}{4pt}
  \resizebox{0.4\textwidth}{!}{
  \begin{tabular}{lccl}
    \Xhline{0.8pt}
    Method & CIDEr & R@1/5 & LLM-AD-eval \\
    \hline
    Video-BLIP2~\cite{Yu_VideoBLIP} & 4.8 & 22.0 & 1.89\ \textbar{} \ - \\
    Video-LLaMA2~\cite{zhang2023video_llama} & 5.2 & 23.6 & 1.91\ \textbar{} \ - \\
    AutoAD-II~\cite{han2023autoad_ii} & 13.5 & 26.1 & 2.08\ \textbar{} \ -  \\
    AutoAD-III~\cite{han2024autoad_iii} & 21.7 & 30.0 & 2.85\ \textbar{} \ - \\
    AutoAD-Zero~\cite{xie2024autoad_zero} & 17.7 & - & 2.83\ \textbar{} \  1.96 \\
    \rowcolor{gray!10} \textbf{DistinctAD} (Ours) & \textbf{22.7}  & \textbf{33.0} & \textbf{2.88}\ \textbar{} \  \textbf{2.03}  \\
    \Xhline{0.8pt}
    \demph{AutoAD-III${\dagger}$}~\cite{han2024autoad_iii} & \demph{25.0} & \demph{31.2} & \demph{2.89\ \textbar{} \  2.01} \\
    \Xhline{0.8pt}
  \end{tabular}
  }
  \vspace{-2mm}
  \caption{Comparisons on CMD-AD. The LLM-AD-eval scores are evaluated with LLaMA2-7B (left) and LLaMA3-8B (right).
  $\dagger$ indicates pre-training on 3.4M HowTo-AD dataset~\cite{miech2019howto100m,han2024autoad_iii}.}
  \label{tab:cmd-sota}
  \vspace{-3mm}
\end{table}

\begin{table}
  \centering
  \setlength{\tabcolsep}{4pt}
  \resizebox{0.4\textwidth}{!}{
  \begin{tabular}{lccc}
    \Xhline{0.8pt}
    Method & CIDEr & R@1/5 & LLM-AD-eval \\
    \hline
    AutoAD-III~\cite{han2024autoad_iii} & 26.1  & - & 2.78\ \textbar{}\ 1.99 \\
    AutoAD-Zero~\cite{xie2024autoad_zero} & 22.6 & 30.6 & \textbf{2.94}\ \textbar{}\ \textbf{2.00} \\
    \rowcolor{gray!10} \textbf{DistinctAD} (Ours) & \textbf{27.4} & \textbf{32.1} & 2.89 \textbar{}\ \textbf{2.00} \\
    \Xhline{0.8pt}
  \end{tabular}}
  \vspace{-2mm}
  \caption{Comparisons on TV-AD. The LLM-AD-eval scores are evaluated using LLaMA2-7B (left) and LLaMA3-8B (right).}
  \label{tab:tvad-sota}
  \vspace{-3mm}
\end{table}

\subsection{Experiment Setup}
\label{subsec:implementation}

\noindent \textbf{Datasets.}
We follow the AD generation benchmark established in AutoAD~\cite{han2023autoad}, conducting experiments on the denoised \textbf{MAD-v2-Named}~\cite{soldan2022MAD} and evaluating on \textbf{MAD-Eval-Named} split. 
Specifically, MAD-v2-Named includes $\sim$330k ADs from 488 movies for training and MAD-Eval has 6,520 ADs crawled from 10 movies for evaluation.
We also evaluate on two recently introduced datasets.
\textbf{CMD-AD}~\cite{han2024autoad_iii} (where ``CMD" stands for Condensed Movie Dataset~\cite{bain2020condensed}) is a movie AD dataset that contains 101k ADs for more than 1432 movies, with 100 movies split for CMD-AD evaluation.
\textbf{TV-AD}~\cite{xie2024autoad_zero} is a recently proposed AD dataset based on TVQA~\cite{lei2018tvqa}, which contains $\sim$31k ADs for training and $\sim$3k ADs for evaluation.

\noindent \textbf{Evaluation Metrics.}
Classic captioning metrics including \textbf{ROUGE-L}~\cite{lin2004rouge}, \textbf{CIDEr}~\cite{vedantam2015cider} and \textbf{SPICE}~\cite{anderson2016spice} are reported to evaluate the quality of generated ADs versus the ground-truth. 
Besides, we also report Recall@\textit{k} within $N$ Neighbours~\cite{han2023autoad_ii} (\textbf{R@k/N}), which calculates the average value of Recall@$k$ for each AD with its $N$ temporally adjacent GT texts, 
where BertScore~\cite{zhang2019bertscore} is used for text similarity matching.
The R@k/N metric is based on retrieving the most similar ground-truth AD among N neighbors, and thus highlights the \textit{distinctiveness} of generated ADs directly.
\textbf{LLM-AD-eval}~\cite{han2024autoad_iii} employs LLMs to assess the quality of generated ADs, assigning scores from 1 (lowest) to 5 (highest). We utilize the LLM prompt from the original study~\cite{xie2024autoad_zero} and apply open-source models LLaMA2-7B-Chat~\cite{touvron2023llama} and LLaMA3-8B-Instruct~\cite{llama3modelcard} for this evaluation.

\noindent \textbf{Implementation Details.}
To facilitate CLIP-AD adaptation in Stage-I, we collect 
the original raw movies from MAD
from platforms like Amazon Prime Video. 
See Appendix \ref{sec:supp_mad_frames} for details.
%
%
We fine-tune the CLIP Vision encoder for 5 epochs with a fixed learning rate 5e-5 using the Adam optimizer~\cite{kingma2014adam} in Stage-I, with a batch size of 512.
In Stage-II, we use a batch of 8 sequences, each containing 16 consecutive video AD-pairs from a movie. 
For each video clip, 8 frames are uniformly sampled.
We use the AdamW~\cite{loshchilov2017adamW} optimizer to train our model for 10 epochs, with a cosine-decayed learning rate and linear warm-up. 
The learning rate is set to $10^{-4}$ for both GPT-2 and LLaMA models.
For external \textit{character} information, we directly use the inference results from AutoAD-Zero~\cite{xie2024autoad_zero} as it gives current best face recognition performance.


\begin{table*}[!htbp]
    \centering
    \small
    \setlength{\tabcolsep}{3pt}
    \begin{subtable}{0.32\textwidth}
        \centering
        \begin{tabular}{ccc}
            \toprule
            Setting & CIDEr & R@5/16 \\
            \hline
            None & 6.7 & 34.0 \\
            Global $\mathcal{L}_g$ & 8.2 & 36.6 \\
            Fine-grained $\mathcal{L}_f$ & 7.7 & 35.2 \\
            \hline
            $\gamma\mathcal{L}_g+ (1-\gamma)\mathcal{L}_f$ &  \textbf{8.6} &  \textbf{36.9} \\
            \bottomrule
        \end{tabular}
        \label{tab:stage_i_a}
        \caption{Stage-I components.}
    \end{subtable}
    \hfill
    \setlength{\tabcolsep}{4pt}
    \begin{subtable}{0.22\textwidth}
        \centering
        \begin{tabular}{ccc}
            \toprule
            Coefficient $\gamma$ & CIDEr  \\
            \hline
            0.1 & 8.0  \\
            0.3 & 8.5 \\
            0.5 & \textbf{8.6} \\
            0.7 & 7.7 \\
            \bottomrule
        \end{tabular}
        \label{tab:stage_i_b}
        \caption{Impact of coefficient $\gamma$.}
    \end{subtable}
    \hfill
    \setlength{\tabcolsep}{4pt}
    \begin{subtable}{0.44\textwidth}
        \centering
        \begin{tabular}{lccc}
            \toprule
            Prompt & Stage-I & CIDEr & R@5/16 \\
            \hline
            \multirow{2}{*}{Contextual ADs~\cite{han2023autoad}} & \ding{55} & 12.6 (17.8) & 39.8 (43.1) \\
            & \ding{51} & \textbf{14.1} (\textbf{19.0}) & \textbf{39.9} (\textbf{44.2}) \\
            \hline
            \multirow{2}{*}{Character~\cite{han2023autoad_ii}} & \ding{55} & 22.0 & 45.6 \\
            & \ding{51} & \textbf{23.1} & \textbf{46.2} \\
            \bottomrule
        \end{tabular}
        \label{tab:stage_i_c}
        \caption{Impact of Stage-I w/ different prompts.}
    \end{subtable}
    
\vspace{-5pt}
\caption{Ablation studies for Stage-I.
(a) Evaluation of global video-AD loss $\mathcal{L}_f$ and fine-grained frame-AD loss $\mathcal{L}_{f}$ on AD performance.
(b) Analysis of the the impact of the coefficient $\gamma$. Both (a) and (b) are conducted with pure visual prompts. (c) Impact of Stage-I when combined with different prompts for the LLM decoder, including contextual ADs and character names. Performance in parentheses indicates results with ground-truth contextual ADs as prompts.}
\label{tab:stage_i}
\vspace{-5mm}
\end{table*}

\subsection{Comparisons with previous methods}
\label{subsec:sota}

We conduct comprehensive comparisons using the widely-adopted MAD-Eval benchmark~\cite{han2023autoad} and two recently introduced AD datasets,  CMD-AD~\cite{han2024autoad_iii} and TV-AD~\cite{xie2024autoad_zero}.

\noindent
\textbf{Comparisons on MAD-Eval} are shown in \cref{tab:mad-sota}. We primarily categorize previous studies into \textit{Training-free} and \textit{Partial-fine-tuning} approaches, as described in \S\ref{sec:literature}.
Our method is a \textit{Partial-fine-tuning} method. 
%
%
When using the same CLIP-B32 and GPT-2, our proposed DistinctAD achieves a CIDEr score of 24.5, surpassing previous AutoAD-I~\cite{han2023autoad} (CIDEr 14.3) and AutoAD-II~\cite{han2023autoad_ii} (CIDEr 19.5).
With our Stage-I adapted CLIP vision encoders
(denoted as CLIP$_{\mathrm{AD}}$), 
we observe stable improvements across all metrics, \eg 25.5 \vs 24.5 on CIDEr and 51.7 \vs 49.8 on recall, validating the effectiveness of our Stage-I strategy.
Notably, DistinctAD with CLIP-AD-B16 and LLaMA3-8B~\cite{llama3modelcard} achieves state-of-the-arts with a CIDEr of 27.3 and Recall@5/16 of 56.0. 
Our outstanding performance on the R@k/N metric demonstrates DistinctAD's ability to generate distinctive ADs, which well match the uniqueness of the clip's contents.
%

Looking at the \emph{training-free} methods, 
despite the capabilities of advanced proprietary VLMs, \eg GPT-4V~\cite{achiam2023gpt4v}, and LLMs, \eg GPT-4~\cite{achiam2023gpt4}, the performance of training-free methods remains inferior to those employing partial-fine-tuning. This discrepancy likely arises from the unique characteristics of AD and movie data, which exhibit a significant domain gap from common vision language training data. As such, these data types were not encountered during the pre-training of proprietary large-scale models.

\noindent
\textbf{Comparisons on CMD-AD and TV-AD.}
We further verify the 
generalizabilty 
 of DistinctAD on the recently proposed CMD-AD  and TV-AD benchmarks, with results presented in Tables \ref{tab:cmd-sota} and \ref{tab:tvad-sota}. For both evaluations, we employ the LLaMA3-8B model.
DistinctAD exhibits superior performance to AutoAD-Zero, AutoAD-II and AutoAD-III in terms of CIDEr and R@1/5 on both CMD-AD and TV-AD.
Meanwhile, DistinctAD exhibits a lower CIDEr compared to AutoAD-III$\dagger$ on CMD-AD, which we conjecture is primarily due to AutoAD-III$\dagger$ pre-training on a very large-scale 3.4M transformed HowTo-AD dataset~\cite{miech2019howto100m,han2024autoad_iii}, \emph{which is currently publicly unavailable.} Despite this, DistinctAD achieves superior R@1/5 performance, underscoring its exceptional ability to generate distinctive and high-quality ADs. This is further corroborated by its leading performance on the LLM-AD-eval metric.

\begin{figure}[t]
  \centering
   \includegraphics[width=1.0\linewidth]{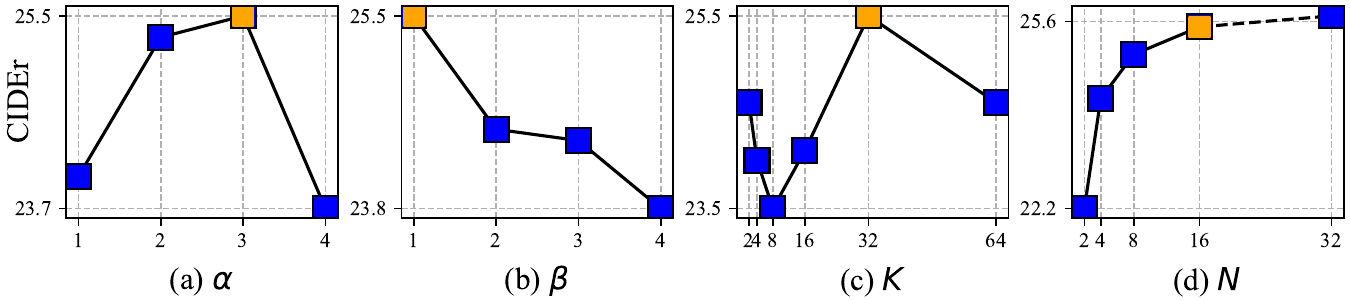}
   \vspace{-19pt}
   \caption{Ablation studies for hyperparameter in Stage-II, with final settings highlighted in \textcolor{orange}{orange}.
   (a) Impact of $\alpha$ on the weight of compact representation $\widehat{\mathcal{H}}$.
   (b) Influence of $\beta$ on cross-attended feature $\widetilde{\mathcal{H}}$.
   (c) Impact of $K$, which denotes the number of clusters in bases $\mathcal{M}$.
   (d) Effect of sampling $N$ consecutive video clips.
   We switch to larger memory GPUs when $N$ exceeds 16.
   } 
   \label{fig:ablations_stage_II}
   \vspace{-3mm}
\end{figure}

\begin{table}[t]
  \centering
  \resizebox{0.4\textwidth}{!}{
  \begin{tabular}{cccccc}
    \toprule
    Ex\#  & $\alpha\widehat{\mathcal{H}}$ & $\beta\widetilde{\mathcal{H}}$ & $\mathcal{L}_{dist}$ & CIDEr & R@5/16 \\
    \hline
    A0 & \ding{55} & \ding{55} & \ding{55} & 23.1 (27.4) & 46.2 \\
    \hline
    B1  & \ding{51} & \ding{55} & \ding{55} & 23.7 (29.3) & 46.6 \\
    B2  & \ding{55} & \ding{51} & \ding{55} & 23.4 (29.1) & 46.1 \\
    B3  & \ding{51} & \ding{51} & \ding{55} & 23.3 (28.1) & 48.0 \\
    \hline
    C1 & \ding{51} & \ding{55} & \ding{51} & 24.3 (29.4) & 50.7 \\
    C2 & \ding{55} & \ding{51} & \ding{51} & 25.3 (\textbf{30.4}) & 51.5 \\
    C3 & \ding{51} & \ding{51} & \ding{51} & \textbf{25.5} (29.8) & \textbf{51.7} \\
    \bottomrule
  \end{tabular}}
  \vspace{-2mm}
  \caption{Ablation studies for components in Stage-II.
  The CIDEr column shows scores with AutoAD-Zero's character~\cite{xie2024autoad_zero} as prompt by default. CIDEr in parentheses represent performance with ground-truth character names.}
  \label{tab:ablation_stage_ii}
  \vspace{-3mm}
\end{table}

\begin{table}[]
    \centering
    \resizebox{0.3\textwidth}{!}{
    \begin{tabular}{cll}
    \toprule
        Consecutive $N$? & CIDEr & R@5/16 \\
        \midrule
        \ding{51} & \textbf{25.5} & 51.7 \\
        \ding{55} & 23.8$_{\textcolor{red}{\downarrow1.7}}$ &  \textbf{52.5}$_{\textcolor{blue}{\uparrow0.8}}$ \\
    \bottomrule
    \end{tabular}}
    \vspace{-2mm}
    \caption{Impact of whether sampling $N$ consecutive clips or not.}
    \label{tab:sequence}
    \vspace{-3mm}
\end{table}

\begin{figure*}[t]
  \centering
   \includegraphics[width=0.96\linewidth]{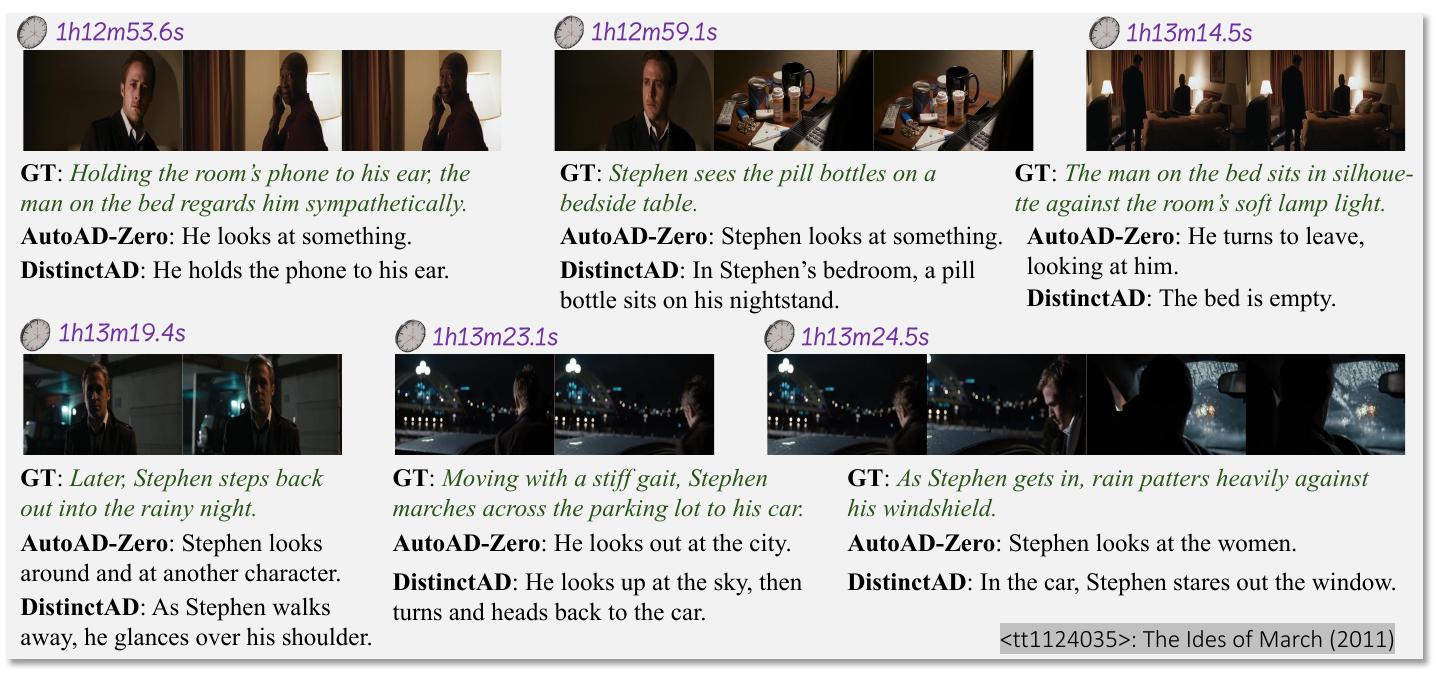}
   \vspace{-6pt}
   \caption{Qualitative results. 
   We present ground-truth (GT) ADs, publicly released AutoAD-Zero outputs, and our DistinctAD predictions for several temporally consecutive movie clips.
   Movie frames are taken from The Ides of March (2011)~\cite{ides_march}. 
   Zoom in for details.
   } 
   \label{fig:visualization}
   \vspace{-3mm}
\end{figure*}

\subsection{Ablation studies}
\label{subsec:ablation}

\textbf{Effect of CLIP-AD Adaptation (I).}
\cref{tab:stage_i}a demonstrates the benefit of our Stage-I strategy, \ie adapting CLIP the vision encoder  to the movie-AD domain via global video-AD matching $\mathcal{L}_g$ and fine-grained frame-AD matching $\mathcal{L}_f$.
In \cref{tab:stage_i}b, 
the balancing coefficient $\gamma$ performs best at 0.5.
In \cref{tab:stage_i}c, our Stage-I strategy consistently enhances performance  when combined with different prompts in the LLM decoder, such as contextual ADs in AutoAD-I~\cite{han2023autoad} or character names in AutoAD-II~\cite{han2023autoad_ii}.
This indicates that our AD-adapted CLIP vision encoder can integrate seamlessly into previous methods, including those training-free models that utilize CLIP-based visual extractors.

\noindent
\textbf{Effect of Distinctive AD narration (II).} 
We evaluate the effectiveness of Stage-II components in \cref{tab:ablation_stage_ii}, based on the default $\mathcal{H}$ (Perceiver's output) and full AD auto-regressive loss $\mathcal{L}_{auto}$. The baseline (A0) outperforms AutoAD-II in CIDEr (23.1 vs. 19.5), primarily due to more accurate character prompts from AutoAD-Zero and the adapted CLIP vision encoder from Stage-I. A0 with AutoAD-II's characters achieves a CIDEr score of 20.6, close to AutoAD-II's performance.
%
Incorporating reconstructed feature $\widehat{\mathcal{H}}$ brings stable improvements on both CIDEr and recall (B1 \& C1), highlighting the importance of compact representations in understanding visual semantics.
Cross-attended feature $\widetilde{\mathcal{H}}$ works better together with distinctive word prediction loss $\mathcal{L}_{dist}$ (C2 \vs B2, C3 \vs B3). We conjecture this is because $\mathcal{L}_{dist}$ provides more definite supervision on re-weighting distinctive words, which guides $\widetilde{\mathcal{H}}$ to attend on concept-related bases.
Overall, applying the full Stage-II pipeline
 brings significant and robust performance (C3).

\noindent
\textbf{Effect of Hyper-parameters.}
\cref{fig:ablations_stage_II} summarizes the ablation studies on 4 hyper-parameters that potentially influence the results in Stage-II.
Coefficient weights $\alpha$ and $\beta$ yield optimal results when set to 3 and 1, respectively. 
This suggests the need to refine our final representations to be more compact for generating ADs.
\cref{fig:ablations_stage_II}(c) shows setting bases number $K$ to 32 yields best. A smaller $K$, \eg 2, can still achieve notable CIDEr, as the Contextual EMA module does not significantly alter the final output $\mathcal{H}_{sum}$.
However, unsuitable values of $K$, \eg 8 or 64, can negatively impact performance.
\cref{fig:ablations_stage_II}(d) reveals that increasing the number of consecutive clips $N$ (with $K$ set to 32) enhances the CIDEr score, though this effect saturates when $N$ exceeds 16.
This demonstrates that more bases should be created to effectively summarize components with additional clips.

\noindent
\textbf{Do the $N$ clips to be consecutive?} 
\cref{tab:sequence} presents the results of sampling \textit{non-consecutive} $N$ clips during training. 
When using non-continuous clips, 
we observe a decrease in the CIDEr metric by 1.7 (25.5 \vs 23.8) because the Contextual EMA module struggles with unrelated contexts. However, the R@5/16 improves by 0.8, which indicates enhancement of the distinctiveness (uniqueness) of the generated AD when using more diverse visual contents. 

\begin{figure}[t]
  \centering
   \includegraphics[width=1.0\linewidth]{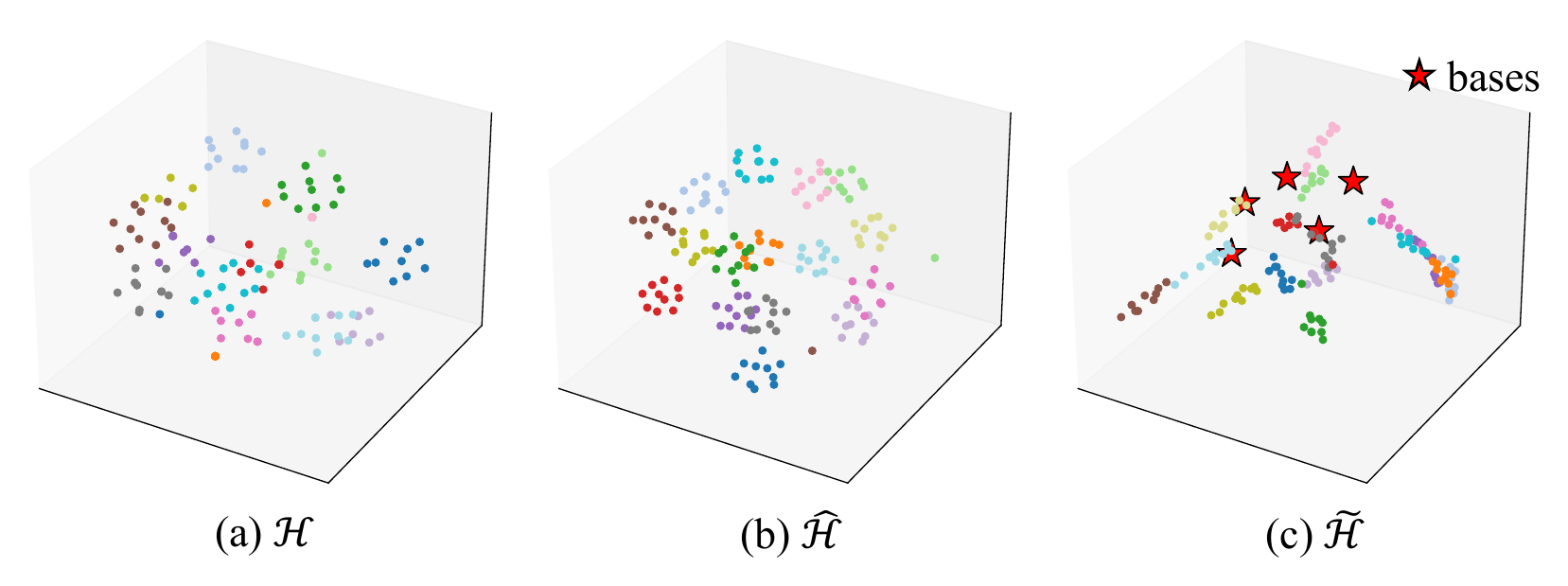}
   \vspace{-25pt}
   \caption{Visualizations of Contextual EMA. 
   (a) A set of randomly generated 3D data $\mathcal{H}$, sampled from $N$ types of samples.
   (b) Compact features $\widehat{\mathcal{H}}$ obtained via Data Re-estimation (DR).
   (c) Cross-attention outputs $\widetilde{\mathcal{H}}$ between $\mathcal{H}$ and bases $\mathcal{M}$.
   } 
   \label{fig:tsne_vis}
   \vspace{-4mm}
\end{figure}

\noindent
\textbf{Visualizations.}
To better understand what Contextual EMA learns, we show the t-SNE~\cite{van2008TSNE} visualizations of $\mathcal{H}, \widehat{\mathcal{H}}$ and $\widetilde{\mathcal{H}}$ (from \S\ref{subsec:stage_2}) in \cref{fig:tsne_vis}, using the same perplexity value across all visualizations.
%
With Contextual EMA, $\widehat{\mathcal{H}}$ exhibits more compact features compared to raw $\mathcal{H}$, \cref{fig:tsne_vis}(b). 
Interestingly in \cref{fig:tsne_vis}(c), cross-attention between $\mathcal{H}$ and bases $\mathcal{M}$ produces strip-like feature distributions pointing to specific base centers, enhancing contextual distinctiveness with improved linear separability and interpretability.

\subsection{Qualitative results.}
\label{subsec:visualization}

\noindent
\cref{fig:visualization} presents qualitative examples of our model. We compare the predictions of DistinctAD (using LLaMA3-8B) with ground-truth captions (GT) and publicly available AutoAD-Zero~\cite{xie2024autoad_zero} outputs.    
Note that clips are sampled \textit{consecutively} in time.
Previous studies often struggle with similar contextual clips, such as those featuring closely-related
scenes and characters, by repeating correct yet insignificant action words, \eg ``look''.
In contrast, 
our DistinctAD effectively generates more engaging ADs by identifying distinctive objects in adjacent clips, \eg ``phone'', ``pill bottle'', and ``car'', along with corresponding more specific behaviors.
More examples can be found in Appendix \S\ref{sec:supp_more_vis}.

\section{Conclusion}
\label{sec:conclusion}

In conclusion, this paper proposes DistinctAD, a novel two-stage framework for generating distinctive audio descriptions for better narrative.
By addressing the domain gap between movie-AD data with a CLIP-AD adaptation strategy, and introducing a Contextual EMA module and a distinctive word prediction loss, our approach significantly improves the quality of AD generation.
The effectiveness of DistinctAD is demonstrated through comprehensive evaluations on multiple benchmark datasets and  ablations studies.
%
Despite these promising results, DistinctAD is still limited by requiring numbers of parameters and the quality of the generated ADs still falls short of human annotations (as reflected by relatively low CIDEr).
%
Overall, automatic AD generation still remains a challenging task, and there is considerable scope for future advancements in this field.

{
    \small
    \bibliographystyle{ieeenat_fullname}
    \bibliography{main}
}
\clearpage
\setcounter{page}{1}
\maketitlesupplementary
\begin{appendices}
\ifappendix 

\section{Analysis of AD Reconstruction with CLIP Embedding Space}
\label{sec:supp_reconstruct_ad}

\begin{figure*}[t]
  \centering
    \includegraphics[width=0.8\linewidth]{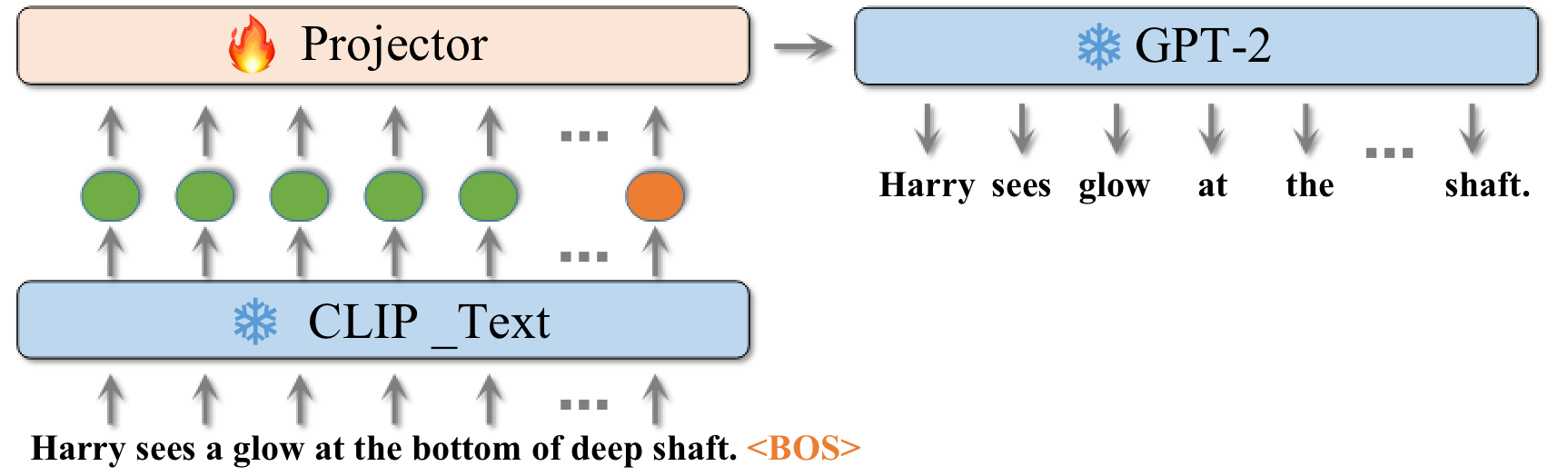}
    \vspace{-10pt}
   \caption{Reconstructing AD words by merely fine-tuning a single-layer projector between a frozen CLIP text encoder and GPT-2. 
   } 
   \label{fig:appen_recons_ad}
    \addtocounter{appendixfigure}{1}
\end{figure*}


\begin{table*}[h]
\small
    \centering
    \begin{tabular}{@{}ccccccccccc@{}}
    \toprule
        \textbf{Projector input} & \textbf{(V)LM }& \textbf{LLM}  & BLEU1 & BLEU2 & BLEU3 & BLEU4 & METEOR & ROUGE-L & CIDEr & SPICE \\
        \midrule
        \texttt{[CLS]} & 
        CLIP-Text & GPT-2
        & 29.3 & 16.4 & 9.2 & 5.1 & 13.2 & 29.4 & 92.2 & 19.4 \\
        \cmidrule(lr){1-11}
        Words &
        CLIP-Text & GPT-2
        & 80.8 & 74.4 & 68.4 & 63.0 & 47.4 & 82.4 & 612.5 & 66.4 \\
    \bottomrule
    \end{tabular}
    \caption{AD reconstruction results on MAD-Eval benchmark. Only textual modality ADs in MAD-Eval are utilized for evaluation, with no movie frames involved. \texttt{[CLS]} denotes using only \textbf{one} class token vector to reconstruct the entire AD.}
    \label{tab:reconstruct_ad_words_supp}
    \addtocounter{appendixtable}{1}
\end{table*}

\noindent
As detailed in the main paper's \S\ref{subsec:stage_1}, our Stage-I strategy, \textbf{CLIP-AD adaption}, is inspired by a preliminary AD reconstruction experiment using the CLIP text encoder~\cite{radford2021CLIP} and GPT-2~\cite{radford2019GPT-2}.
We begin with the question: \emph{is the CLIP text embedding space expressive enough for embedded AD words to be reconstructed by LLMs?} 
If the reconstruction process is successful---meaning that LLMs can understand the textual ADs encoded by the CLIP text encoder---then the misalignment in the VLM joint feature space likely occurs because of the CLIP vision encoder, 
rather than between the CLIP text encoder and the LLMs.
On the other hand, 
if the reconstruction is not successful, then the pre-trained CLIP joint embedding space is not suitable for the AD task, and both text and vision encoders need to be retrained.

To address this question, we design the AD words reconstruction pipeline illustrated in \cref{fig:appen_recons_ad}. 
Specifically, we input the AD sentence into a frozen CLIP text encoder, modified to output tokens for each word. 
We implement two versions of AD reconstruction: 1) using \textbf{only a single \texttt{[CLS]} vector}, or 2) using \textbf{all word tokens} as prompts.
We append a \texttt{<BOS>} tag to signal the start of reconstruction.
The output embeddings are then fed into a learnable single-layer projector, transforming the CLIP word tokens into the LLM embedding space.
We apply an auto-regression loss identical to (\ref{eq:auto_regress}) in the main paper, with the visual \texttt{prompt} setting as none.
The projector is trained for 10 epochs on MAD-v2-Named~\cite{soldan2022MAD} ADs, and the performance is evaluated using classical n-gram based metrics on the MAD-Eval benchmark~\cite{han2023autoad}.
The reconstruction results are presented in \cref{tab:reconstruct_ad_words_supp}.
Remarkably, by merely fine-tuning a single-layer projector, AD reconstruction achieves results closely aligned with the ground truth, such as scores of \textbf{80.8} on \textbf{BLEU1} and \textbf{612.5} on \textbf{CIDEr} with all words input. 
Additionally, using only a single \texttt{[CLS]} vector to recover the entire AD achieves 92.2 on CIDEr, \emph{significantly} outperforming existing AD works, which score around 20 CIDEr.
This demonstrates that AD words (or [CLS] vector) encoded by the CLIP text encoder can be effectively understood by LLMs, suggesting that the misalignment mainly lies within the joint VLM feature space, \ie, discrepancies between CLIP vision embeddings and CLIP AD embeddings.

\section{Analysis of Neighboring (Contextual) Features}
\label{sec:supp_similar_ctx}

\begin{figure*}[t]
  \centering
\includegraphics[width=0.85\linewidth]{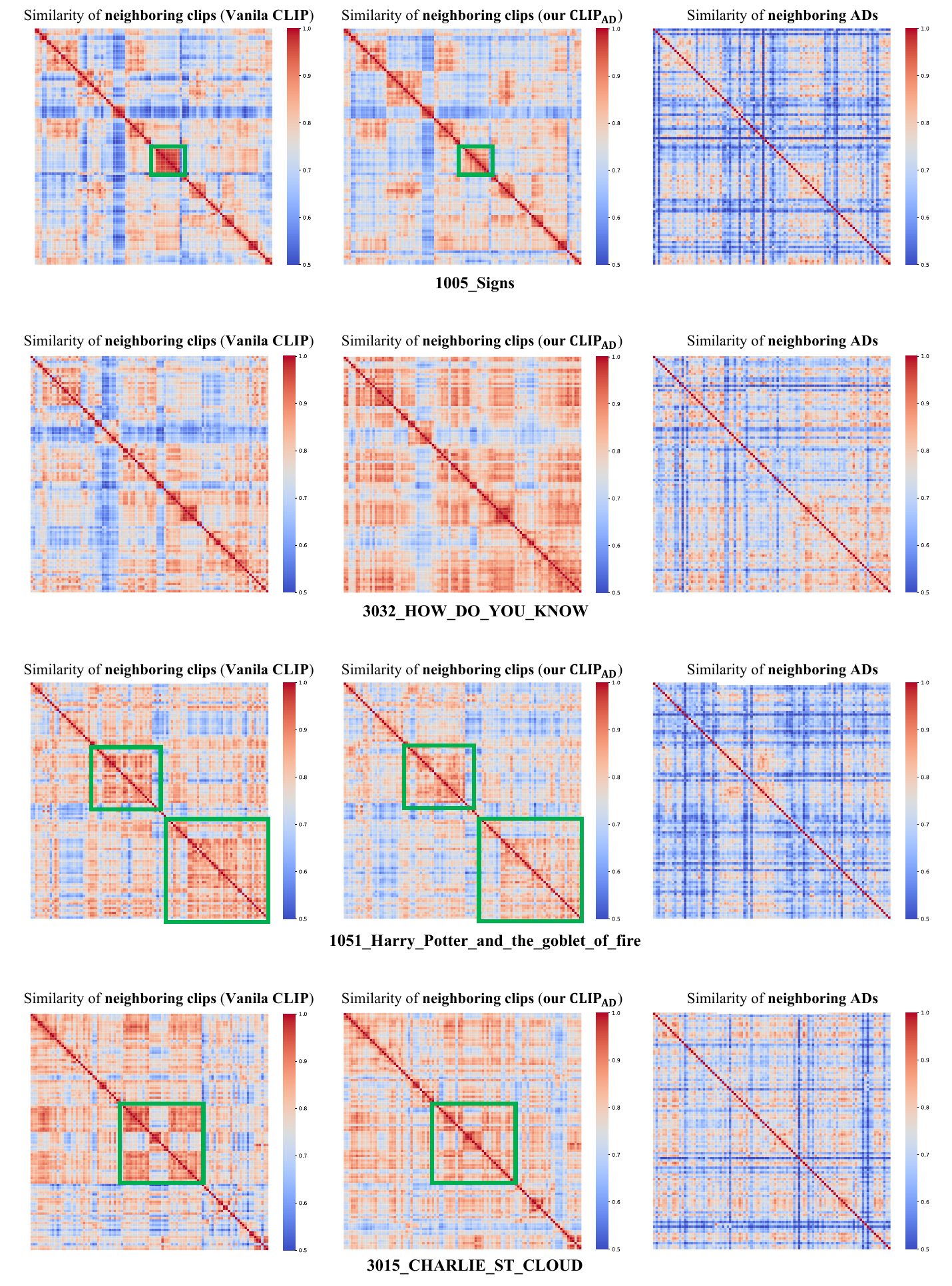}
   \caption{
   Cosine similarity matrices of neighboring (contextual) movie clips using vanilla CLIP (left) and our adapted CLIP$_{\mathrm{AD}}$ in Stage-I (middle).
   We also show similarity matrices of corresponding neighboring ADs (right).
   Movie clips are from Signs (2002), How Do You Know (2010), Harry Potter and the Goblet of Fire (2005), and Charlie St. Cloud (2010). Green boxes indicate differences between vanilla CLIP and our CLIP-AD. Zoom in for details. 
   } 
   \label{fig:appen_context_sim}
    \addtocounter{appendixfigure}{1}
\end{figure*}

In this section, we validate our primary hypothesis:
\textit{sequential clips from an extended video often share redundant scenes or characters, resulting in similar visual features within contexts,} as discussed in \S\ref{subsec:stage_2} of the main paper. \cref{fig:appen_context_sim} presents the cosine similarity matrix for neighboring (contextual) movie clips (left) and their corresponding audio descriptions (ADs) (right) from four randomly selected films. 
The visual clip features are derived through mean pooling over $T$ frame embeddings encoded by the CLIP Vision encoder, while the AD features are obtained from the \texttt{[CLS]} embeddings encoded by the CLIP Text encoder.
From these similarity matrices, we observe two key points:
(i) Movie clips generally exhibit greater similarity to each other compared to ADs, indicated by a higher proportion of red (deep) colors; 
(ii) Compared to ADs, neighboring (contextual) movie clips show prominent areas of similarity around the diagonals (i.e., the block diagonal structure), demonstrating that they share similar visual features due to recurring scenes and characters.

In \cref{fig:appen_context_sim}, middle column, we illustrate the similarity of neighboring movie clips using our adapted CLIP$_{\mathrm{AD}}$ vision encoder in Stage-I (see \S\ref{subsec:stage_1} of the main paper).
Significant changes compared to \textit{vanilla CLIP} visualizations are highlighted with \textcolor{green}{green} rectangles.
Our CLIP$_{\mathrm{AD}}$ helps reduce redundancy among neighboring video clips, as evidenced by the smaller similarity values within the green rectangles, which helps to improve the generation of distinctive ADs in our framework. 
This further demonstrates the effectiveness of our Stage-I strategy.

\section{Detailed Formulation of CrossAttention}
\label{sec:supp_x_attn}
In this part, we provide an in-depth explanation of the Cross-Attention formulation, building upon (\ref{eq:x_attn}) in the main paper. The query $Q$ originates from the Perceiver output, denoted as $\mathcal{H}$, while both the key $K$ and the value $V$ are derived from the base matrix $\mathcal{M}$.
We apply three Linear layers to transform the query, key, and value into a unified embedding space, as represented by the following equations:
\begin{align}
    Q &= \mathcal{H} W_{Q}^{T} + b_{Q}, \\
    K &= \mathcal{M} W_{K}^{T} + b_{K}, \\
    V &= \mathcal{M} W_{Q}^{T} + b_{V}.
\end{align}
Subsequently, the cross-attention mechanism is formulated by computing a weighted sum of the values, where the weights are determined by the similarity between the queries and keys. The softmax function ensures the normalization of the attention weights. The final cross-attention output $\widetilde{\mathcal{H}}$ is given by:
\begin{equation}
    \widetilde{\mathcal{H}} = \mathrm{Softmax}(\frac{QK^T}{\sqrt{d_k}})V,
\end{equation}
where $\sqrt{d_k}$ acts as a scaling factor to stabilize the gradient flow during training.

\section{Additional Qualitative Examples}
\label{sec:supp_more_vis}
Following \cref{fig:visualization} in the main paper, we present additional qualitative examples in \cref{fig:appen_more_vis}, utilizing our adapted CLIP-AD-B16~\cite{radford2021CLIP} in Stage-I and LLM LLaMA3-8B~\cite{llama3modelcard}. 
The movie clips are \textbf{consecutively} sampled from the following films: \textbf{(a) }\textit{Signs} (2002), \textbf{(b)} \textit{The Roommate} (2011), and \textbf{(c)} \textit{How Do You Know} (2010), listed from top to bottom. For accurate retrieval and alignment, the starting time of each movie clip is indicated in the top-left corner of each clip. Additionally, we provide results from the publicly available AutoAD-Zero~\cite{xie2024autoad_zero} for comparison. 
The numerous high-quality examples further demonstrate the superiority of our proposed method, DistinctAD.

Since complete predictions and codes are \textit{unavailable} for many previous methods, such as AutoAD-I, AutoAD-II, AutoAD-III, and MM-Narrator, we only collect the qualitative examples presented in their original papers and perform qualitative comparisons in \cref{fig:appen_single_vis}. 
Training-free methods are highlighted with a blue background, while partial-fine-tuning methods are marked in orange.
 It is evident that training-free methods utilizing proprietary models like GPT-4 or GPT-4V often encounter hallucination issues, producing irrelevant or imaginary details.
In contrast, partial-fine-tuning methods, \ie AutoAD-I, AutoAD-II and DistinctAD, generate more accurate ADs close to human-annotated ground-truth. (We use past 3 \textit{ground-truth} ADs as AutoAD-I's textual prompts.)
Despite this, AutoAD-I can be negatively influenced by its contextual content, \eg ``nuns" mistakenly appears in \textbf{(d)}.
AutoAD-II tends to generates similar AD words, \eg \textit{``furrowed brow"} for movie frames with close-up faces in \textbf{(a)} and \textbf{(d)}, whereas our DistinctAD is generally more distinctive.

\begin{figure*}[h]
  \centering
   \includegraphics[width=1.0\linewidth]{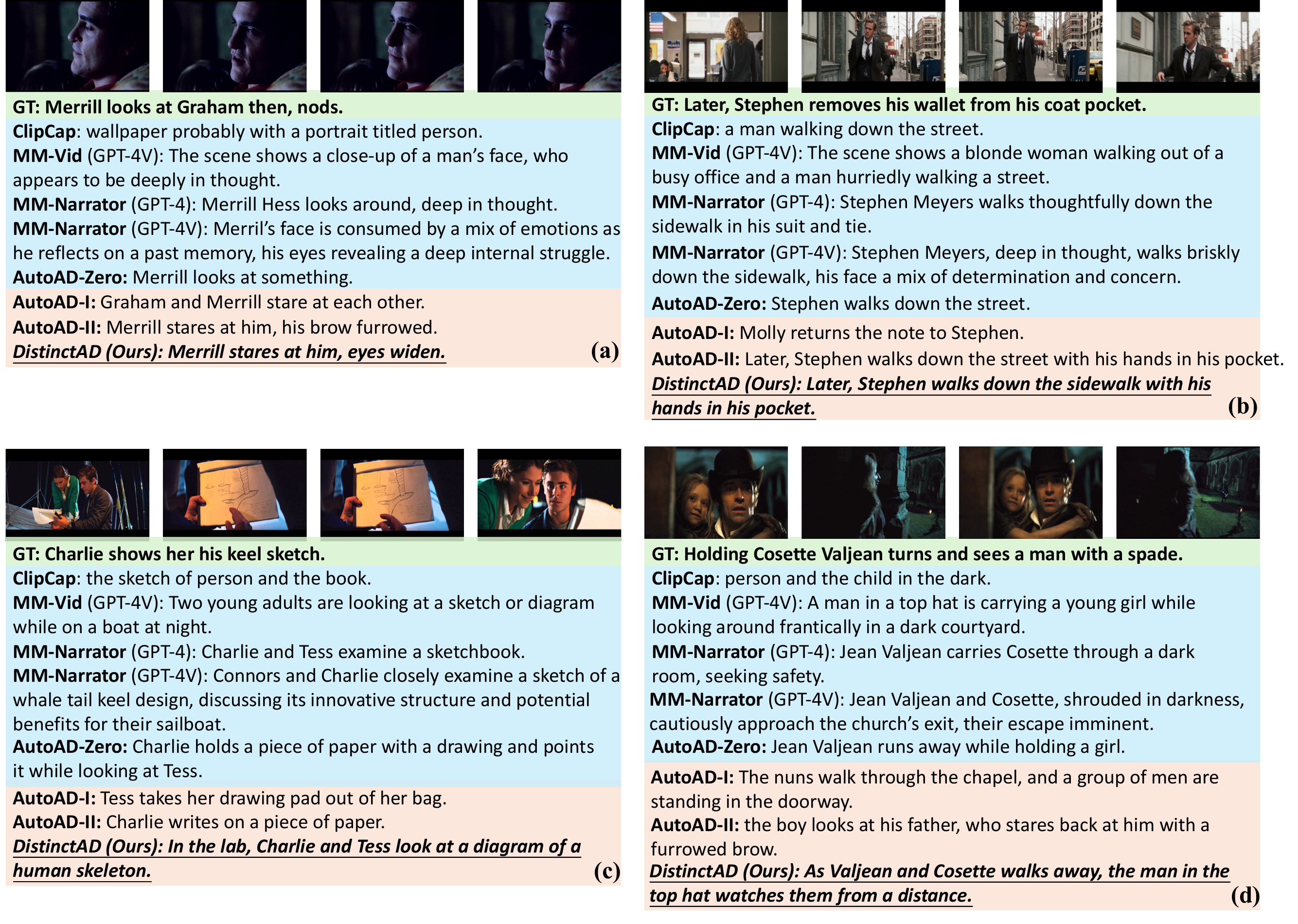}
   \vspace{-20pt}
   \caption{Qualitative comparisons on \textbf{single} movie clips between ClipCap, MM-Vid, MM-Narrator, AutoAD-Zero, AutoAD-I, AutoAD-II, and our DistinctAD. The movies are from \textbf{(a)} Signs (2002), \textbf{(b)} Ides of March (2011), \textbf{(c)} Charlie St. Cloud (2010), and \textbf{(d)} Les Miserables (2012).
   Zoom in for details.
   } 
   \label{fig:appen_single_vis}
    \addtocounter{appendixfigure}{1}
\end{figure*}

\begin{figure*}[h]
  \centering
   \includegraphics[width=0.85\linewidth]{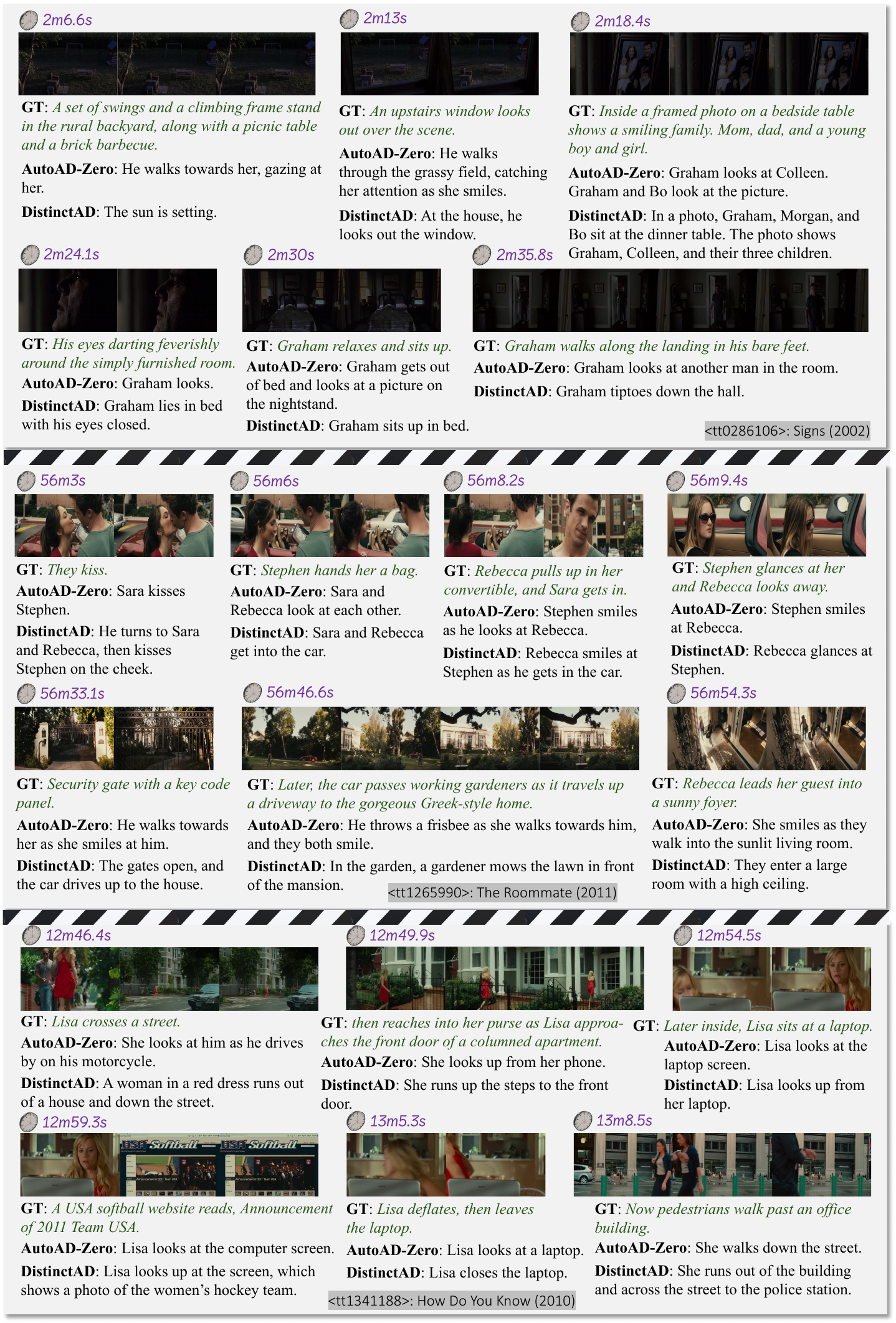}
   \vspace{-10pt}
   \caption{More qualitative results on \textbf{consecutive} movie clips. 
   Movie frames from top to bottom are taken from Signs (2002), The Roommate (2011), How Do You Know (2010), respectively. 
   Zoom in for details.
   } 
   \label{fig:appen_more_vis}
    \addtocounter{appendixfigure}{1}
\end{figure*}


%

\section{Raw Frames of MAD}
\label{sec:supp_mad_frames}
Due to copyright restrictions, MAD~\cite{soldan2022MAD} only provides frame-level movie features extracted by CLIP~\cite{radford2021CLIP}.
However, to facilitate CLIP-AD adaptation in Stage-I, we require raw MAD movie frames to fine-tune the CLIP vision encoder.
To achieve this, we collect MAD raw movies from third-party platforms such as Amazon Prime Video.
Out of the 488 movies in the MAD-train list, 3 are not available online, as shown in \cref{tab:3_missing_video_supp}.

\begin{table}[t]
    \centering
    \begin{tabular}{rrr}
    \toprule
        MAD\_ID & IMDB\_ID & Movie Title \\
        \midrule
        \texttt{4797} & \texttt{tt0395571} & \textit{Holy Flying Circus} \\
        \texttt{4839} & \texttt{tt4846340} & \textit{Halo: The Fall of Reach} \\
        \texttt{5900} & \texttt{tt0408306} & \textit{Murdered by My Father} \\
    \bottomrule
    \end{tabular}
    \caption{Meta information of missing films in MAD-train.}
    \label{tab:3_missing_video_supp}
    \addtocounter{appendixtable}{1}
\end{table}

\noindent
Moreover, due to geographical differences, we may download different versions of movies, potentially leading to mismatches between movie clips and annotated timestamps. 
To address this, we conduct a thorough check by comparing our downloaded movies with the MAD dataset and their metadata in the IMDB database.
Out of 488 movies, 9 have time durations that vary more than one minute. Details are shown in \cref{tab:temporal_varied_movie}.

\begin{table}[t]
    \centering
    \footnotesize
    \begin{tabular}{@{}rrrrr@{}}
    \toprule
        MAD\_ID & IMDB\_ID & MAD\_Time & Our\_Time & IMDB\_Time \\
        \midrule
        \texttt{2738} & \texttt{tt0450232} & 1h 37m 26s & \textcolor{green}{1h 41m 59s} & 1h 42m \\
        \texttt{2787} & \texttt{tt1136608} & 1h 19m 24s & \textcolor{green}{1h 52m 16s} & 1h 52m \\
        \texttt{4017} & \texttt{tt5463162} & \textcolor{green}{1h 59m 20s} & 1h 57m 41s & 1h 59m \\
        \texttt{4061} & \texttt{tt1837636} & 1h 28m 2s & \textcolor{green}{2h 8m 12s} & 2h 8m \\
        \texttt{4266} & \texttt{tt0375735} & 1h 36m 8s & \textcolor{green}{1h 40m 39s} & 1h 40m \\
        \texttt{4772} & \texttt{tt0424136} & 1h 39m 53s & \textcolor{green}{1h 44m 33s} & 1h 44m \\
        \texttt{4902} & \texttt{tt0119310} & \textcolor{green}{1h 15m 30s} & 1h 11m 55s & 1h 14m \\
        \texttt{5634} & \texttt{tt2929690} & \textcolor{green}{1h 40m 52s} & 1h 51m 50s & 1h 40m \\
        \texttt{6952} & \texttt{tt2527338} & 2h 31m 52s & \textcolor{green}{2h 21m 53s} & 2h 21m \\
    \bottomrule
    \end{tabular}
    \caption{Metadata for movies with duration difference exceeding 1 minute. Durations closer to the IMDB time are highlighted in \textcolor{green}{green}. }
    \label{tab:temporal_varied_movie}
    \addtocounter{appendixtable}{1}
\end{table}

\noindent
According to the statistical information in \cref{tab:temporal_varied_movie}, we identify potential temporal misalignment noise in the existing MAD benchmark.
To mitigate negative impacts during training, we exclude movies with durations that significantly differ from those in the IMDB database.
The removed movie IDs are: \texttt{4017, 4902, 5634.}
A summary of the final employed MAD-v2-Named training dataset is provided in  \cref{tab:denoised_mad}.

\begin{table}[t]
    \centering
    \begin{tabular}{crr}
    \toprule
        MAD-v2-Named & \# movies & \# AD \\
        \midrule
        MAD-Train-Features \cite{soldan2022MAD} & 488 & 334,296 \\
        MAD-Train-Frames (Ours) & 482 & 326,632\\
    \bottomrule
    \end{tabular}
    \caption{Statistics of our refined MAD dataset incorporating raw frames.}
    \label{tab:denoised_mad}
    \addtocounter{appendixtable}{1}
\end{table}

\end{appendices}


\end{document}